\documentclass{article}

\usepackage{microtype}
\usepackage{graphicx}
\usepackage{subfigure}
\usepackage{booktabs} 

\usepackage{hyperref}

\usepackage[accepted]{icml2025}

\usepackage{amsmath}
\usepackage{amssymb}
\usepackage{mathtools}
\usepackage{amsthm}

\usepackage{float}
\usepackage{adjustbox}      
\usepackage{bm}             
\usepackage{multirow}       
\usepackage[symbol]{footmisc}   
\usepackage{tabularx}
\definecolor{tabfirst}{rgb}{1, 0.7, 0.7} 
\definecolor{tabsecond}{rgb}{1, 0.85, 0.7} 
\definecolor{tabthird}{rgb}{1, 1, 0.7} 

\usepackage[capitalize,noabbrev]{cleveref}

\theoremstyle{plain}

\theoremstyle{definition}

\theoremstyle{remark}

\def\transp{\mathsf{T}}

\usepackage[textsize=tiny]{todonotes}

\icmltitlerunning{sshELF: Single-Shot Hierarchical Extrapolation of Latent Features}

\begin{document}

\twocolumn[
\icmltitle{sshELF: Single-Shot Hierarchical Extrapolation of Latent Features \\ for 3D Reconstruction from Sparse-Views}



\icmlsetsymbol{equal}{*}

\begin{icmlauthorlist}
\icmlauthor{Eyvaz Najafli}{equal,con,tue}
\icmlauthor{Marius Kästingschäfer}{equal,con,fre}
\icmlauthor{Sebastian Bernhard}{con}
\icmlauthor{Thomas Brox}{fre}
\icmlauthor{Andreas Geiger}{tue}
\end{icmlauthorlist}

\author{%
  \textbf{Eyvaz Najafli}$^{1,3,*}$
  \quad
  \textbf{Marius Kästingschäfer}$^{1,2,*}$ 
  \quad
  \textbf{Sebastian Bernhard}$^{1}$\\
  \quad
  \textbf{Thomas Brox}$^{2}$
  \quad
  \textbf{Andreas Geiger}$^{3}$\\
  $^1$Continental
  \quad
  $^2$University of Freiburg
  \quad
  $^3$University of Tübingen\\
  \texttt{eyvaz.najafli@student.uni-tuebingen.de}\\
  \texttt{marius.kaestingschaefer@continental.com}
}

\icmlaffiliation{con}{Continental}
\icmlaffiliation{tue}{University of Tübingen}
\icmlaffiliation{fre}{University of Freiburg}

\icmlcorrespondingauthor{Eyvaz Najafli}{eyvaz.najafli@student.uni-tuebingen.de}
\icmlcorrespondingauthor{Marius Kästingschäfer}{marius.kaestingschaefer@continental.com}

\icmlkeywords{Computer Vision, Few-View-to-3D, Autonomous Driving}

\vskip 0.3in
]



\printAffiliationsAndNotice{\icmlEqualContribution} 

\begin{abstract}
Reconstructing unbounded outdoor scenes from sparse outward-facing views poses significant challenges due to minimal view overlap. 
Previous methods often lack cross-scene understanding and their primitive-centric formulations overload local features to compensate for missing global context, resulting in blurriness in unseen parts of the scene. 
We propose sshELF, a fast, single-shot pipeline for sparse-view 3D scene reconstruction via hierarchal extrapolation of latent features. 
Our key insights is that disentangling information extrapolation from primitive decoding allows efficient transfer of structural patterns across training scenes.
Our method: (1) learns cross-scene priors to generate intermediate virtual views to extrapolate to unobserved regions, (2) offers a two-stage network design separating virtual view generation from 3D primitive decoding for efficient training and modular model design, and (3) integrates a pre-trained foundation model for joint inference of latent features and texture, improving scene understanding and generalization. 
sshELF can reconstruct 360$^{\circ}$ scenes from six sparse input views and achieves competitive results on synthetic and real-world datasets. 
We find that sshELF faithfully reconstructs occluded regions, supports real-time rendering, and provides rich latent features for downstream applications. The code will be released.
\end{abstract}

\section{Introduction}
\label{sec:intro}

\begin{figure}
    \centering
    \includegraphics[width=1\linewidth]{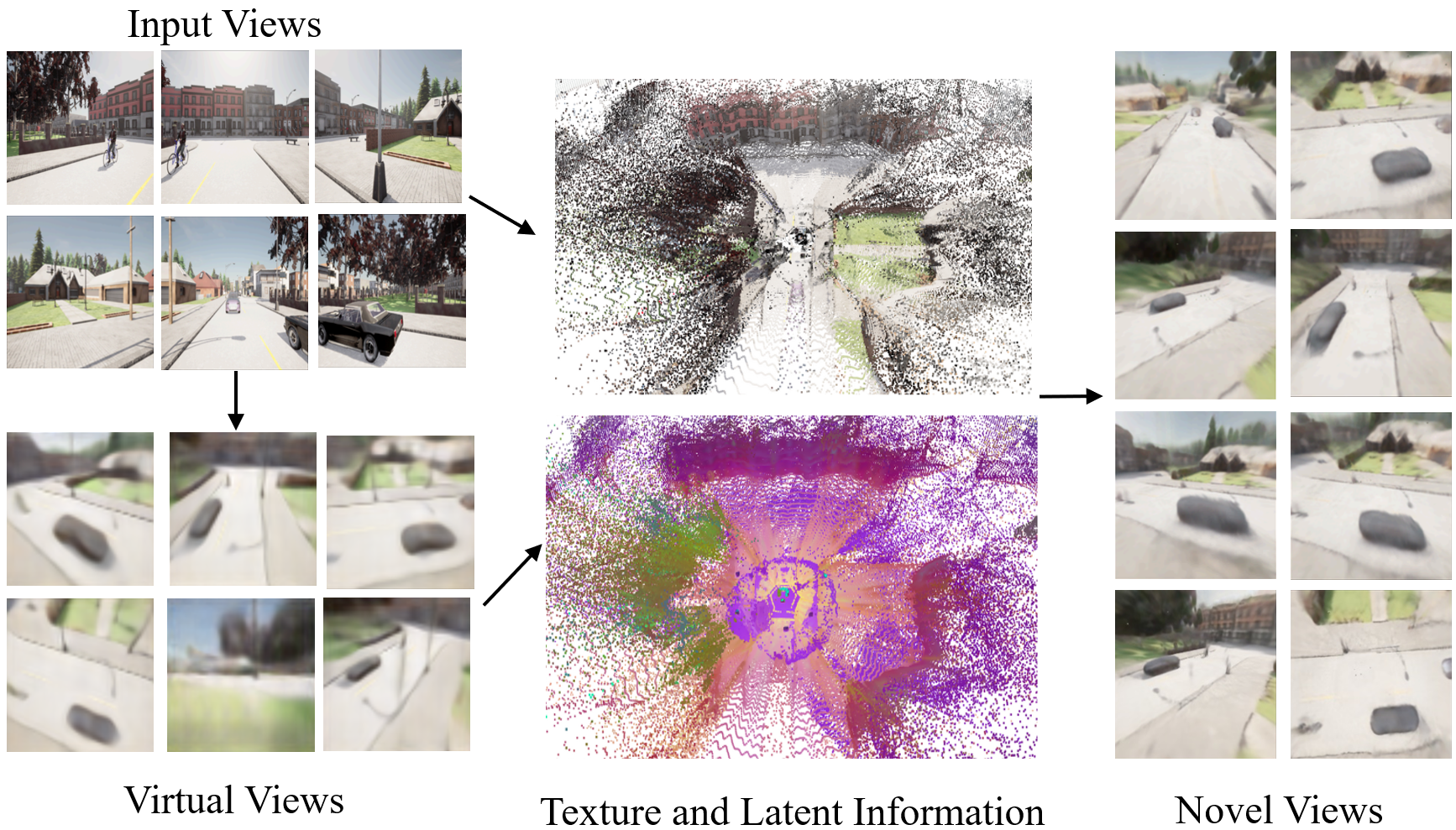}
    \caption{\textbf{Overview}. Given a number of input images, sshELF first reconstructs several virtual views and only then predicts the 3D Gaussian primitives of the scene from which novel views are rendered. The colors of the latent information correspond to different object classes, such as purple for buildings and green for vegetation. 
    \vspace{-0.4cm}}
    \label{fig:teaser}
\end{figure}

We consider the problem of reconstructing unbounded outdoor scenes from sparse outward-facing cameras with very little overlap between adjacent views. Solving this problem poses two fundamental challenges: (1) resolving distant object occlusions (areas hidden behind terrain or other vehicles) and ego-occlusions (regions obscured by the sensor platform itself, for example below the vehicle), and (2) overcoming limited multi-view correspondence cues due to minimal overlap. In practical autonomous driving applications, this dual challenge becomes even more demanding since dense bird's-eye views or occupancy maps are required in real-time.



While traditional neural radiance fields (NeRFs) and 3D Gaussian Splatting have advanced novel view synthesis, their reliance on per-scene optimization and dense view coverage limits applicability in real-world scenarios with sparse, non-overlapping inputs \cite{mildenhall2020nerf, kerbl3Dgaussians}.  Existing per-scene optimization methods are often tested on inward-facing datasets with high view overlap \cite{realestate10K, reizenstein2021commonobjects3dlargescale, deitke2022objaverseuniverseannotated3d} or where novel views are close to the input views, simplifying cross-view triangulation. In contrast, vehicle-mounted cameras are usually outward-facing with minimal camera overlap \cite{behley_semantickitti_2019, caesar_nuscenes_2020, Sun_2020_CVPR} and operate in large, unbounded outdoor environments. Recent feedforward approaches aim to generalize across scenes but struggle with large viewpoint changes \cite{MVSplat, yu2020pixelnerf}, some lack support for multi-view aggregation necessary for 360-degree surround-view synthesis, \cite{szymanowicz2024flash3d}, or are not real-time renderable \cite{gieruc20246imgto3d}.

Although Vision transformers trained on large datasets for metric depth prediction can provide useful priors \cite{bhat2023zoedepth, yin2023metric3d, marigold_ke2023repurposing}, the resulting depth maps, when combined with pixel information, are inadequate for generating complete 3D representations \cite{surround_monodepth2022, yang2024scaleawaresurroundmonodepthtransformers, SEED4D}. A key limitation is their inability to account for unobserved regions, including areas occluded by terrain or other objects and those obscured by the sensor platform itself, such as the ground beneath the vehicle. As a result, subsequent reconstructions exhibit incomplete geometry in those regions. Furthermore, when reconstructing geometry from multiple depth maps simultaneously, multi-view scale inconsistency at the border regions leads to artifacts. Thus resulting in low-quality 3D reconstructions and inadequate novel views \cite{szymanowicz2024flash3d, SEED4D}. \\
\vspace{-0.4cm}

\begin{figure} 
    \centering
    \includegraphics[width=1\linewidth]{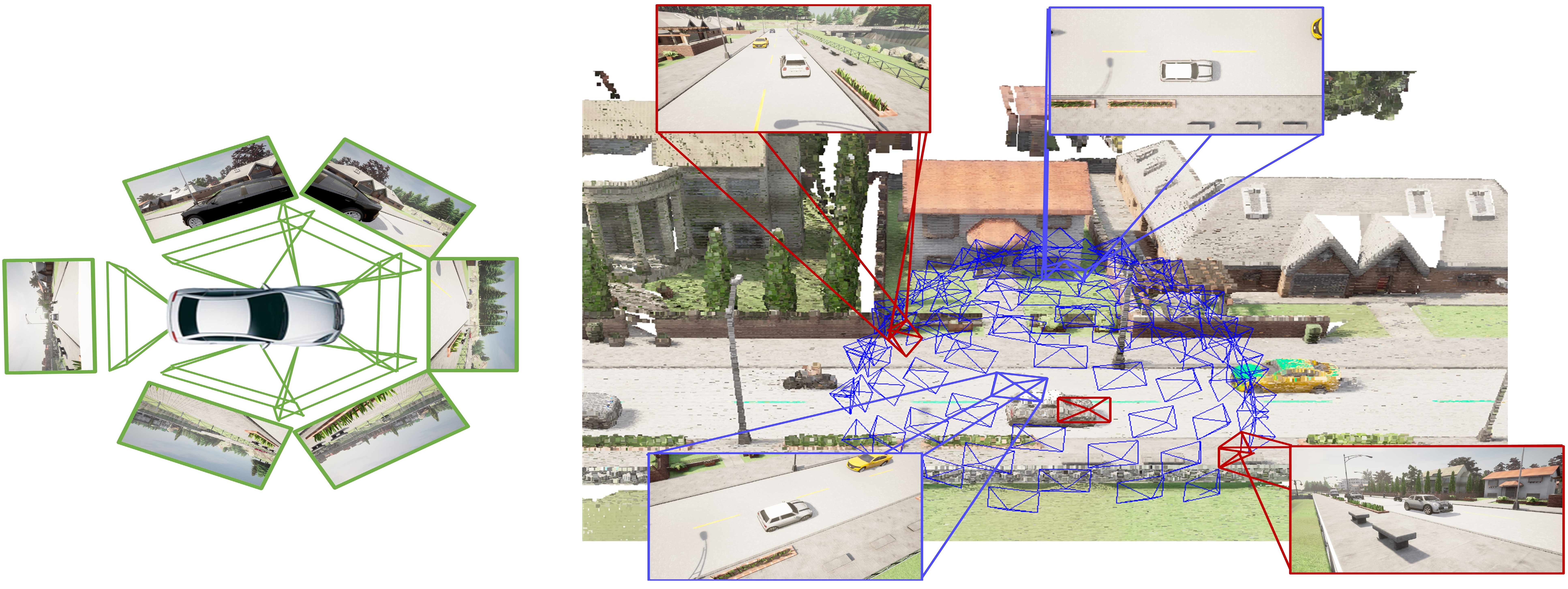}
    \caption{\textbf{Reference, Virtual and Novel Views}. An example showing input views in green, a set of virtual views in red, and potential novel views in blue. Virtual view generation is key to enhancing representational capacity and extrapolating to unobserved scene areas.
    \vspace{-0.4cm}}
    \label{fig:virtual_views}
\end{figure}

\noindent \textbf{Problem Statment}. \label{subsec:problem_statement} Our work tackles the problem of fast single-shot sparse-view 3D reconstruction for outdoor traffic scenes. We introduce an efficient, performant \underline{s}ingle-\underline{s}hot pipeline for sparse-view 3D reconstruction via \underline{h}ierarchical \underline{e}xtrapolation of \underline{l}atent \underline{f}eatures called \textbf{sshELF}. The paper is based on three key insights. 

First, existing models are constrained in their ability to infer unseen regions and views far from the input images since their representational capacity is limited \cite{MVSplat, pixelsplat_Charatan_2024_CVPR}. By restricting the process only to the information present in the input images, without intermediate steps or representations, the models struggle to generalize to unobserved parts of the scene. Unlike previous methods, sshELF generates several intermediate virtual views that help to reconstruct unseen regions. This way, our method is not restricted to the information in the given input images. In figure \ref{fig:virtual_views} we visualize example input, virtual and novel views.

Second, our network is decomposed into a \textit{backbone} that generates virtual views and a \textit{translator} that decodes the reference and virtual views into explicit Gaussian primitives from which novel views can be rendered in real-time. The separation enables the backbone to increase the information content while allowing the translator to lift higher-quality Gaussian primitives. As a byproduct, the decomposition facilitates isolated training of the two stages, leading to a significant reduction in computational requirements. Unlike wide-baseline volumetric representation \cite{LaRa}, our virtual views do not require dense sampling and have a lower memory overhead. Compared to end-to-end training, the decomposed approach further increases both the number and resolution of virtual views the backbone can generate. This thereby increases the amount of information one can pass to the translator. The partitioning into the backbone and translator further makes the design process more flexible since the sub-networks can be trained independently.

Third, large pre-trained foundation models such as DinoV2 \cite{oquab2023dinov2} and near-metric depth estimation methods \cite{depth_anything_v2} are underutilized when employed solely as input to the model \cite{szymanowicz2024flash3d}. Unlike previous works, we incorporate pre-trained feature extractor and depth estimation models directly into our architecture. Incorporating pre-trained models as fundamental building blocks enables our backbone to output both texture and latent information for reference and virtual views. Our results outperform previous methods, suggesting that the rich training signal plays a crucial role. Properly leveraging the intermediate latent features of a pre-trained foundation model subsequently enables the optimal use of depth prediction models relying on multi-stage latent features. Moreover, obtaining latent information alongside Gaussian primitives opens up potential downstream applications such as semantic scene understanding or 3D detection. See Figure \ref{fig:teaser} for an example latent clustering.

Overall, sshELF is a fast two-stage single-shot unbounded 3D scene reconstruction pipeline, particularly well suited for outward-facing cameras with little view overlap. The contributions of this paper can be summarized as follows:
\textbf{(1) Reconstruct Occluded Regions}. Our method reconstructs occluded regions faithfully where other methods fail, as shown by competitive results on both synthetic and real-world data.
\textbf{(2) High Speed and Visual Quality}. Our method can perform fast end-to-end 360$^{\circ}$ scene reconstruction and novel view synthesis from six unbounded surround vehicle views in 0.18s. sshELF can render far-away viewpoints with high quality as measured on synthetic and real-world data.
\textbf{(3) Optimal Utilization of Latent Information}. Since our method jointly predicts latent and texture information, as well as depth, we can leverage latent features to obtain insights into spatial semantics, occupancy, and geometry.

\noindent We perform an in-depth analysis to justify our architectural choices and compare our final model with multiple state-of-the-art approaches. We will make our code available.

\section{Related Work}
\label{sec:background}

\textbf{Iterative Driving Scene Reconstruction}. Iterative driving scene reconstruction methods perform test-time per-scene fitting by incrementally updating the scene representation until reaching a predefined step count or convergence criteria. This makes them infeasible for real-time applications compared to feed-forward methods. Iterative scene reconstruction methods can broadly be classified into NeRF-based and 3D Gaussian-based approaches. 

An early NeRF-based method, Neural Scene Graph (NSG)~\cite{NSG_2021_CVPR} decomposes a scene into static and dynamic objects, representing their relation via a hierarchical directed graph. Since then, a number of methods have used scene graphs or decompositions into static, dynamic, and flow fields to model scenes \cite{ml_nsg_2024_CVPR, NeuRAD_Tonderski_2024_CVPR, turki2023suds, yang2023emernerf}. These methods often require LiDAR data and are slow to render, with training times exceeding 30 minutes in some cases. MARS~\cite{wu2023mars} similarly uses a foreground-background decomposition, while READ~\cite{li2022read} and StreetSurf~\cite{guo2023streetsurf} focus only on static scenes.

Gaussian-based parameterizations have gained popularity due to their ability to perform real-time rendering. Many of them model backgrounds and objects separately, using bounding boxes to identify the objects. Several Gaussian-based approaches also use scene graphs~\cite{Hugs_Zhou_2024_CVPR, yan2024street, chen2024omnire, zhou2024drivinggaussian}. To alleviate artifacts per-scene reconstruction pipelines have been combined with diffusion-based priors~\cite{VEGS, Yu2024SGDSV} or by enforcing symmetry~\cite{khan2024autosplatconstrainedgaussiansplatting}. Most per-scene optimization methods rely on posed input images, though a few methods also jointly learn poses~\cite{chen2023periodic, li2024_VDG}. Despite being real-time renderable, offline reconstruction remains time-intensive and often LiDAR-dependent, limiting real-time applicability. \\

\begin{figure*}[t!]
    \centering
    \includegraphics[width=1\textwidth]{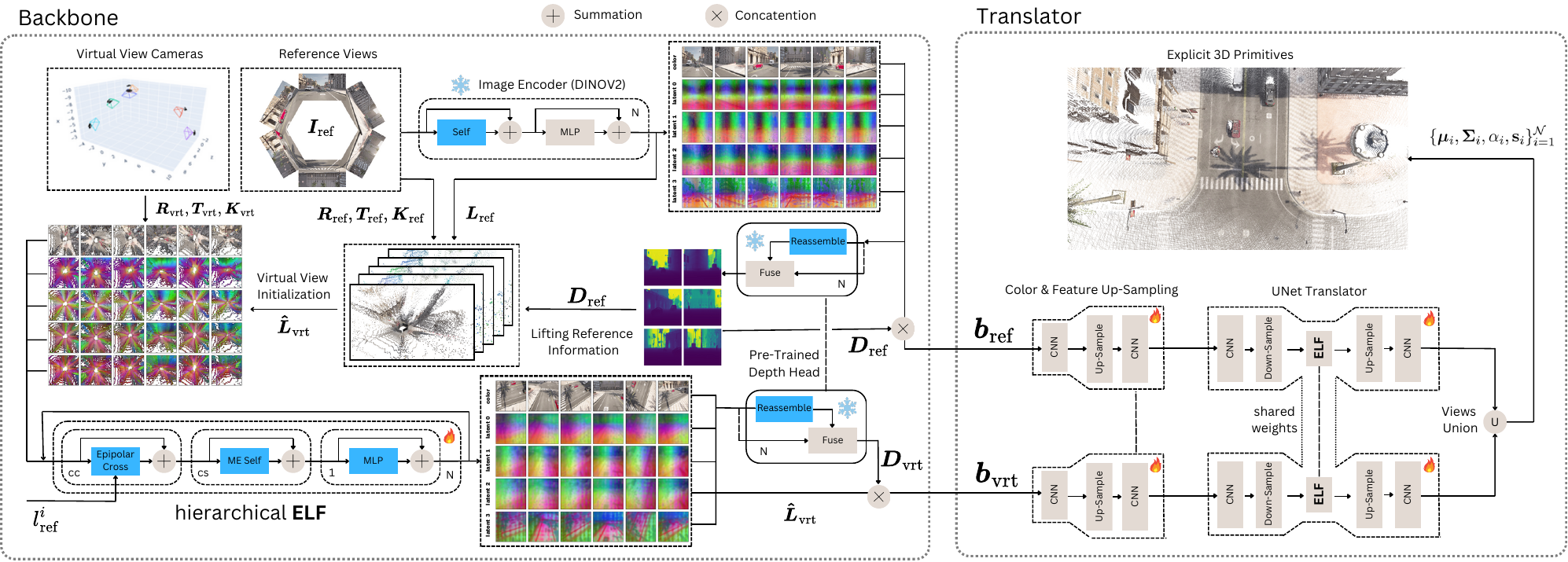}
    \caption{\textbf{Overview of sshELF}. Given a few input images, sshELF first encodes them into latent features using a pre-trained DinoV2 (Sec.\ref{image_encoder}). As part of the \textit{backbone}, the latent features, together with a pre-trained depth head, are used to initialize the virtual views, which are refined using hierarchical ELF blocks consisting of cross- and self-attention layers (Sec. \ref{backbone}). Reference and virtual views are then fed into the \textit{translator} part to predict 3D Gaussian splats (Sec. \ref{translator}). Not shown here is the rasterization part used for creating novel views (Sec. \ref{rendering_nvs}).}
    \label{fig:sshELF_model}
\end{figure*}

\noindent \textbf{Few-View Reconstruction}. Iterative per-scene optimization is time-consuming and constrained in its generalizability, prompting the development of feedforward methods. 

Early NeRF-based methods retrieve image features via view projection and aggregate the resulting features~\cite{yu2020pixelnerf, wang2021ibrnet, mvsnerf}. Some methods apply additional constraints using diffusion priors~\cite{ReconFusion_Wu_2024_CVPR} or time-consuming iterative diffusion-based refinement~\cite{szymanowicz2023viewset_diffusion, RenderDiffusion_Anciukevicius_2023_CVPR}. Many of the mentioned papers focus on small-scale scenes or single objects. A method focusing on single-shot prediction in driving scenarios is DistillNeRF~\cite{wang2024distillnerf}, which distills single-shot priors from the per-scene optimization method EmerNeRF. Closest to our work are Neo360~\cite{irshad2023neo360}, which is limited to inward-facing views, and 6Img-to-3D~\cite{gieruc20246imgto3d}, which uses a slow triplane-based representation.

Gaussian-based methods explicitly model scenes and enable a simpler single-shot parameterization compared to neural rendering approaches. Recent works focus on single objects or small-scale scenes~\cite{xu2024grm, TriplaneGaussian_Zou_2024_CVPR, yang2024gaussianobject, szymanowicz24splatter}, but require input views with significant overlap. Flash3D~\cite{szymanowicz2024flash3d} uses depth prediction but fails to inpaint unseen regions. Methods like pixelSplat~\cite{pixelsplat_Charatan_2024_CVPR}, latentSplat~\cite{wewer2024latentsplat}, and MVSplat~\cite{MVSplat} leverage cross-attention for image pairs, excelling in close-range novel view synthesis but struggling with large camera displacements. Concurrent work DrivingForward~\cite{tian2024drivingforwardfeedforward3dgaussian} reconstructs driving scenes from nuScenes but is limited to small viewpoint changes between consecutive timeframes.

Many few-shot 3D reconstructions and novel view synthesis methods overload the 3D Gaussian predictor to inpaint occluded parts of the scene by predicting several Gaussians per ray. This causes blurriness in the unseen parts of the scene, which are far away from the input views. Unlike existing methods, our work utilizes intermediate representations to generate unobserved views and thus obtain a more complete scene reconstruction.

\section{Method}
\label{sec:methods}

We distinguish between reference, virtual, and novel views. Reference views describe the captured viewpoints fed as input into the architecture, and novel views describe the novel synthesized viewpoints. While previous work focuses on reference and novel views, we additionally introduce virtual views that define intermediate viewpoints between them, facilitating the inpainting of unobserved regions in the final reconstruction. 

Given $n_{\text{ref}}$ reference views containing RGB images $\bm{I}_{\text{ref}} \in \mathbb{R}^{3\times H\times W}$, their associated camera extrinsics $\bm{P}_{\text{ref}} = \left[\bm{R}_{\text{ref}} | \bm{T}_{\text{ref}} \right] \in \mathbb{R}^{3\times4}$ and intrinsics $\bm{K}_{\text{ref}} \in \mathbb{R}^{3\times3}$, sshELF generates consistent 3D geometry and synthesize $n_{\text{nvs}}$ novel surround views $\bm{I}_{\text{nvs}} \in \mathbb{R}^{3\times H\times W}$. The method is visualized in Figure~\ref{fig:sshELF_model}. The number of reference, virtual, and novel views can be flexibly varied within our pipeline. The image encoder, backbone, translator, and rendering process are described in the following sections.

\subsection{Image Encoder}\label{image_encoder}

Given images $\bm{I}_{\text{ref}}$, sshELF applies a pre-trained self-supervised vision transformer (ViT) \cite{dosovitskiy2020vit} to obtain patch-wise latent features $\bm{L}_{\text{ref}}=\{\boldsymbol{l}^{i}_{\text{ref}} \mid i=1\text{,...,} n_{\text{ref}} \} \in{\mathbb{R}^{n_{\text{ref}} \times (d_E + 3) \times (H / 14) \times (W / 14)}}$, and a class token [CLS] $\{\boldsymbol{g}^{i}_{\text{ref}}\}^{n}_{i=1}\in{\mathbb{R}^{d_E}}$ containing aggregated global feature information.  The model retrieves latent embeddings from the last $n$ ViT blocks at various depths, each of which is a $d_E$ dimensional vector. We use DINOv2~\cite{oquab2023dinov2} since it is semantically rich and retains geometric information well. To optimally preserve texture information, the normalized and resized RGB tensor is concatenated with each of the $n$ layer's patch embeddings, increasing the channel dimension of $\boldsymbol{l}^{i}_{\text{ref}}$ to $d_E + 3$.

\subsection{Backbone: Generating Virtual Views}\label{backbone}

Given the intrinsics $\bm{K}_{\text{ref}}$ and extrinsic $\bm{P}_{\text{ref}}$ of the reference views, along with ground-truth image information and intrinsics $\bm{K}_{\text{vrt}}$ and extrinsic $\bm{P}_{\text{vrt}}$ of $n_{\text{vrt}}$ virtual views, the task of the backbone is to construct latent and texture information for the virtual views. During cross-scene training we randomly sample the virtual views with probabilities proportional to the degree of overlap with the reference views to maximize the information content of occluded regions. 

To initialize $n_{\text{vrt}}$ virtual views, texture information from the reference views is projected into 3D and back-projected onto the virtual views using point cloud rendering~\cite{ravi2020pytorch3d}. To obtain the required depth maps $\bm{D}_i$, the latents $\bm{L}_{\text{ref}}$ without texture information obtained from the reference views are fed into a fine-tuned dense depth prediction transformer~\cite{DPT_Ranftl_2021_ICCV}. The resulting depth maps $\bm{D_i}$ can then be used to project pixel information from the reference views into 3D space by
\vspace{-0.0cm}
\begin{equation} \label{equation:reference_feature_lift}
    \mathbf{M_i} = \begin{bmatrix}
    \mathbf{R}_{\text{ref}} & \mathbf{T}_{\text{ref}} \\
    \mathbf{0}^\transp & 1
  \end{bmatrix}^{-1}
  \begin{bmatrix}
    \mathbf{K}_{\text{ref}} & \mathbf{0} \\
    \mathbf{0}^\transp & 1
  \end{bmatrix}^{-1} \text{ } \mathbf{\hat{m}_{i}}\text{,}
\end{equation}
where $\mathbf{\hat{m}_{i}} = \begin{bmatrix}\mathbf{m_{i}} & 1 & 1/\mathbf{D}_{i} \end{bmatrix}^\transp$ representing the pixel coordinate in image space and $M_i$ representing the 3D point in world coordinates. In the next step, the feature information from the reference views is projected into 3D space to initialize the latent and texture information of the virtual views $\tilde{\bm{L}}_{\text{vrt}}=\{\boldsymbol{l}^{i}_{\text{vrt}} \mid i=1\text{,...,}n \}$. The initialized virtual views are now more informative than randomly initialized ones since they contain approximate texture and latent information, but they still suffer from occlusions and other artifacts. To refine the virtual views, we hierarchically \underline{e}xtrapolate the \underline{l}atent \underline{f}eatures via \textbf{ELF} blocks, each denoted as $\mathcal{\bm{E}(\cdot)}$. Ground Truth reference view information, virtual view initializations, and the output from the previous blocks are fed as input to the corresponding ELF block. In each ELF block, a succession of $cc$ cross-attention denoted as \textbf{CA} and $cs$ self-attention denotes as \textbf{SA} operations are applied to transfer information between reference and virtual views, followed by an \textbf{MLP} layer:
\vspace{-0.2cm}
\begin{align}
    \begin{split}\label{equation:local_elf_iterations_1}
    \boldsymbol{\hat{l}}^{i}_{\text{vrt}} \text{  } &= \boldsymbol{\hat{l}}^{i-1}_{\text{vrt}} + \boldsymbol{\tilde{l}}^{i}_{\text{vrt}}  
    \end{split}\\
    \begin{split}\label{equation:local_elf_iterations_2}
    C(\boldsymbol{\hat{l}}^{i}_{\text{ref}},  \boldsymbol{\hat{l}}^{i}_{\text{vrt}})_{j} + &= \textbf{CA}(C(\boldsymbol{\hat{l}}^{i}_{\text{ref}}, \boldsymbol{\hat{l}}^{i}_{\text{vrt}})_{j}) \text{ }\text{ } \forall j \in\{1, \ldots, c c\}
    \end{split}\\
    \begin{split}\label{equation:local_elf_iterations_3}
    C(\boldsymbol{\hat{l}}^{i}_{\text{ref}}, \boldsymbol{\hat{l}}^{i}_{\text{vrt}})_{k} + &=\textbf{SA}(C(\boldsymbol{\hat{l}}^{i}_{\text{ref}}, \boldsymbol{\hat{l}}^{i}_{\text{vrt}})_{k}) \text{ }\text{ }  \forall k \in\{1, \ldots, c s\}
    \end{split}\\
    \begin{split}\label{equation:local_elf_iterations_4}
    C(\boldsymbol{\hat{l}}^{i}_{\text{ref}}, \boldsymbol{\hat{l}}^{i}_{\text{vrt}}) +&= \textbf{MLP}(C(\boldsymbol{\hat{l}}^{i}_{\text{ref}}, \boldsymbol{\hat{l}}^{i}_{\text{vrt}}))\text{,}
    \end{split}
\end{align}
where $C(\cdot\text{, }\cdot)$ concatenates two feature maps along the view dimension. The ELF block performs an inpainting-like refinement, addressing occlusions and enhancing texture. By predicting features for both virtual and reference views, we introduce a cycle-consistency constraint, ensuring that the ELF block preserves the reference features as close to the ground truth as possible while reconstructing virtual views. Following~\cite{wewer2024latentsplat}, we use epipolar geometry to inform and constrain the cross-attention computation described in equation \eqref{equation:local_elf_iterations_2}. A sequence of CNN and MLP layers is applied on the output of equation \ref{equation:local_elf_iterations_4} to infer the \texttt{[CLS]} token from the latent features.

In the final stage of the backbone, the resulting multi-stage latent codes for the virtual views are passed through a pre-trained depth head to obtain near-metric depth maps $D_{vrt}$. To achieve a complete scene reconstruction, we concatenate latent, texture, and depth predictions for the virtual views with the ground truth information from the reference views and feed them all as input to the translator. 

\subsection{Translator: Lifting to 3D}\label{translator}

The translator transforms the features resulting from the backbone into 3D Gaussian primitives. The input consists of aggregated latent, color, and approximate depth information of both reference and virtual views.
\vspace{-0.1cm}

\begin{equation} \label{definition:backbone_output}
    C(\boldsymbol{b}_{\text{ref}}\text{, }\boldsymbol{b}_{\text{vrt}})  \in {\mathbb{R}^{(n_{\text{ref}} + k_{\text{vrt}}) \times (d_E + 3 + 1) \times (h) \times (w)}}
\end{equation}

where $h \times w$ is the spatial dimension of the latent embeddings. We utilize a UNet-like architecture \cite{unet_2015, song2021scorebased} to map from each view's backbone features to 3D Gaussian splats. The translator architecture $\mathcal{T}(\cdot)$ consists of two parts: an encoder and a decoder. The output of equation \ref{definition:backbone_output} is fed into the encoder, which is passed through a single layer of \textbf{ELF} block $ \mathcal{\bm{E}}$ to enforce multi-view consistency over reference and virtual views on the lowest layers in the UNet. The decoder then predicts splats per view whereby intermediate skip connections between the decoder and the encoder help to preserve fine-grained details. The resulting Gaussian splats are projected to ego vehicle coordinates, concatenated across spatial and views dimensions. The scene is now parameterized with the 3D Gaussian primitives
\vspace{-0.1cm}

\begin{equation} \label{equation:gaussian_predict}
    \{\bm{\mu}_i, \bm{\Sigma}_{i}, \alpha_{i}, \mathbf{s}_{i}\}^{(n_{\text{ref}} + n_{\text{vrt}}) \times N}_{i=1} 
\end{equation}
where $\bm{\mu}_{i}\in\mathbb{R}^{3}$ is the per pixel location of the Gaussians, $\bm{\Sigma}_{i}\in\mathbb{R}^{3\times 3}$ is the covariance, $\alpha_{i} \in [0, 1)$ the opacity and $\mathbf{s}_{i}\in\mathbb{R}^{(l+1)^2}$ represents the coefficients for the spherical harmonics of degree $l$ and $N$ stands for the number of Gaussians generated per view. To ensure that the covariance matrix remains positive semi-definite, we predict a diagonal scaling matrix $\mathbf{S}$ and orthonormal rotation matrix $\mathbf{R}$ such that $\mathbf{\Sigma}=\mathbf{R} \mathbf{S} \mathbf{S}^\transp \mathbf{R}^\transp$, parameterized by the axis-scales $\mathbf{s} \in\mathbb{R}^{3}$ and $\mathbf{q} \in\mathbb{R}^{4}$ defining a normalized quaternion \cite{kerbl3Dgaussians}. To alleviate local minima during the learning process of 3D primitives, we apply a probabilistic depth map prediction similar to pixelSplat \cite{pixelsplat_Charatan_2024_CVPR}.

\subsection{Rendering Novel Views}\label{rendering_nvs}

Given extrinsics $\bm{P}_{\text{nvs}}$ and intrinsics $\bm{K}_{nvs}$, we render novel views $\bm{\hat{I}}_{\text{nvs}}$ using gaussian rasterization. To compute the unnormalized density of the $i^{th}$ 3D Gaussian the following function is applied 
\vspace{-0.0cm}
\begin{equation} \label{equation:gaussian_unnormalized_density}
    G_{i}(\mathbf{x}) = \text{exp}\Bigl( -\frac{1}{2}(\mathbf{x} - \bm{\mu}_i)^\transp \mathbf{\Sigma}^{-1}_{i} (\mathbf{x} - \bm{\mu}_i) \Bigl).
\end{equation}

The color of the Gaussians $\bm{c}\in\mathbb{R}^3$ when viewed from direction $\bm{d}\in\mathbb{R}^3$ is computed by summing the spherical harmonics basis $\bm{c}(\bm{d}) = \sum_{i=1}^{n} s_{i} \mathcal{B}_i (\bm{d})$, here $ \mathcal{B}_i$ is the $i^{\mathrm{th}}$ spherical harmonics basis function. Finally, the pixel intensity $\bm{c}$ is computed from the $(n_{\text{ref}} + n_{\text{vrt}}) \times N$ ordered Gaussians using alpha compositing in the following way
\vspace{-0.1cm}

\begin{equation}
\boldsymbol{c}=\sum_{i=1}^n \boldsymbol{c}_i \alpha_i \prod_{j=1}^{i-1}\left(1-\alpha_j\right)
\end{equation}

The Gaussians can be rendered in real-time following \cite{kerbl3Dgaussians}. The full architecture remains end-to-end differentiable; however, we train the backbone and the translator separately to be able to increase the number and resolution of inferred virtual views. 

\subsection{Training Objectives}

During cross-scene training, the model is compelled to learn transferable structural priors, enabling generalization across different scenes. We assume the virtual views to be available for supervision, which can be sampled from the existing data in practice. This allows us to obtain the ground truth virtual view features by feeding them through the pre-trained ViT blocks to get $\{\boldsymbol{l}^{i}_{\text{vrt}}\}_{i=1}^n$ for the training of the backbone. The translator can be trained separately with ground truth reference and virtual views. \\

\noindent \textbf{Backbone Objectives}. The backbone reconstructs virtual views consisting of latent features and texture information from reference views. An MSE loss between reconstructed features and ground truth features is applied at every stage of the backbone:
\vspace{-0.0cm}
\begin{equation}
    \mathcal{L}_{bb} = 
    \lambda_1 \mathcal{L}_{\text{MSE}}\left(\boldsymbol{\hat{L}}_{\text{vrt}}, \boldsymbol{L}_{\text{vrt}} \right) + 
    \lambda_2 \mathcal{L}_{\text{MSE}}\left(\boldsymbol{\hat{L}}_{\textbf{ref}}, \boldsymbol{L}_{\text{ref}} \right) 
\end{equation}
where $\lambda_1$, $\lambda_2 > 0$. The first part of the loss enforces the backbone to reconstruct intermediate and final latent features and low-resolution texture information of the virtual views. As an additional constraint, the second part of the loss enforces processed reference features to be similar to the ground truth reference features. This can be viewed as a cycle-consistency constraint whereby the information flowing from reference views to virtual views and then back should be maintained with minimal change. \\

\noindent  \textbf{Translator Objectives}. The translator predicts the Gaussian primitives from which the novel views are rasterized. The translator is trained with:
\vspace{-0.0cm}
\begin{equation}
    \mathcal{L}_{tr} = 
     \lambda_3 \mathcal{L}_{\text{MSE}}\left(\boldsymbol{ \hat{I}}_{\text{nvs}}, \boldsymbol{I}_{\text{nvs}} \right) + 
    \lambda_4 \mathcal{L}_{\text{MAE}}\left(\boldsymbol{\hat{D}}_{\text{ref}}, \boldsymbol{D}_{\text{nvs}} \right) 
\end{equation}
where $\lambda_3>0$, $\lambda_4 \geq 0$ and $\boldsymbol{\hat{D}}_{\text{ref}}$ is the $Z$-buffer (also known as depth buffer) retrieved from the renderer, which provides a depth approximation \cite{kerbl3Dgaussians}. This way, the error enforces texture and geometric correspondence where ground-truth depth maps are available, in the case of nuScenes $\lambda_4$ is set to zero. When point cloud information is available, we also experimented with adding a Chamfer distance loss.

\section{Experiments}
\label{sec:results}

A number of experiments are conducted to assess the capabilities of our method. We use synthetic and real-world driving data (Section \ref{subsec:datasets}) to test the performance of our method and consistency of our results across datasets. Our experimental setting highlights implementation details and applied metrics (Section \ref{subsec:experiment_settings}). Our quantitative and qualitative results (Section \ref{subsec:results}) show the visual quality and the high inference speed of our method. Our analysis and ablations (Section \ref{subsec:ablation}) demonstrate the necessity of the different model components.

\subsection{Datasets}\label{subsec:datasets}

\textbf{SEED4D}. We make use of the publically available Synthetic Ego--Exo Dynamic 4D (SEED4D) dataset \cite{SEED4D}. The dataset consists of synthetic ego- and exocentric views. The static version of the dataset that we use for training contains a total of 212k images from 2k driving scenes. Per scene six outward-facing ego-vehicle images and 100 spherical images for supervision exist. The ego-centric camera setup resembles the relative camera placement within the nuScene dataset, whereby the overlap between adjacent outward-looking views is minimal. We use the ego-vehicle images as reference views and sample virtual and novel views from the exocentric views. We follow the default split and use Town 1, 3 to 7, and 10 for training and reserving 100 scenes from Town 2 for testing. This configuration provides 1900 locations for cross-scene training. \newline

\noindent \textbf{NuScenes}. We additionally test on the well-established nuScenes dataset \cite{caesar_nuscenes_2020}. It comprises six ego vehicle views with only 10\% view overlap \cite{tian2024drivingforwardfeedforward3dgaussian} from 1000 driving scenes of 20 seconds. Each sequence comprises around 240 frames per camera, of which 40 are keyframes. We only use the keyframes during our experiments and follow the default split of 700 scenes for training and 150 for testing.
Since NuScenes does not contain exocentric views, we construct a multi-view evaluation setup by aggregating egocentric views captured across temporal sequences. In our framework, we utilize views with a temporal difference (TD) of zero (i.e., simultaneous captures from all vehicle-mounted cameras) as reference views. Novel view synthesis targets are then defined at TD=2, 3, and 4 timesteps after reference timestep, equivalent to 1s, 1.5s, and 2s temporal offsets respectively. Virtual views are place between reference and novel views.

\subsection{Experimental Setup}\label{subsec:experiment_settings}

\textbf{Baselines}. We compare our method against a number of methods from the original SEED4D paper and additional recent few-image novel-view-synthesis baselines. PixelNeRF~\cite{yu2020pixelnerf} uses projected image feature for conditioning a neural radiance field. SplatterImage~\cite{szymanowicz24splatter} predicts pixel-aligned 3D Gaussian primitives using a U-net.  MVSplat~\cite{MVSplat} utilizes cross-attention, a cost volume, and a pre-trained depth model. 6Img-to-3D~\cite{gieruc20246imgto3d} uses self- and cross-attention for parameterizing a triplane together with image feature projection. PixelSplat~\cite{pixelsplat_Charatan_2024_CVPR} utilizes epipolar cross-attention and performs a probabilistic prediction of pixel-aligned Gaussians. For nuScenes we focus exclusively on the most recent real-time capable methods.   \\

\noindent \textbf{Implementation Details}. We implement sshELF using PyTorch \cite{PyTorch_NeurIPS2019}, a memory-efficient attention mechanism \cite{xFormers2022} and the renderer implementation from the original 3DGS paper \cite{kerbl3Dgaussians}. Each hierarchical ELF block consists of two (cc) epipolar-cross attention parts and one (cs) self-attention block. The total number of ELF blocks is four in the backbone and one in the translator.

During cross-scene training, we set the number of reference views to 6, the number of virtual views to 6, and the number of novel views to 2. 
For the SEED4D dataset, the input views have a resolution of $896 \times 896$, the resolution of the virtual views in the backbone is $64 \times 64$, the DINO features are obtained with an input resolution of $896 \times 896$, and novel views have a resolution of $256 \times 256$.
When working with the nuScenes dataset, most values remain the same except for the rendering resolution of sshELF, which is increased to $896 \times 896$. The same resolutions are used for SplatterImage, pixelSplat, and MVSplat.
 
The translator is trained on ground truth reference and virtual views until convergence. Once the backbone is converged, the translator is fine-tuned with the estimated virtual views until convergence. We leverage the knowledge learned over the synthetic dataset and continue training on the nuScene dataset from the best model checkpoints for 100K steps. During the training of the backbone $\lambda_1 = 1000.0$, $\lambda_2 = 0.1$ and for the translator training $\lambda_3 = 100.0$ and $\lambda_4 = 0.001$. 
We use an Adam optimizer \cite{Kingma_Adam} and an intial learning rate of $1 \times 10^{-5}$, cosine annealing.
The backbone is trained using an A40 GPU with 48 GB, and the translator using a V100 GPU with 32 GB.\\

\noindent \textbf{Evaluation Metrics}. Performance is measured using the peak signal-to-noise ratio (PSNR), structural similarity index (SSIM) \cite{ssim}, and learned perceptual image patch similarity (LPIPS) \cite{zhang2018unreasonable}. We additionally compute the depth root mean square error (D-RMSE) where ground truth metric depth is available or the Chamfer distance when LiDAR data is accessible.

\subsection{Results}\label{subsec:results}

\textbf{SEED4D}. As shown in Table~\ref{tab:results_seed4d}, sshELF outperforms all previous methods in terms of PSNR, and ranks second in SSIM and D-RMSE. Other methods suffer from incomplete geometry, particularly in hidden or sensor-blocked regions. sshELF achieves a runtime of 0.182 seconds, demonstrating competitive performance. Notably, sshELF's end-to-end performance is more than 15x faster than the second-best model, 6Img-to-3D.
\begin{table}[!ht]
    \centering
    
    \setlength{\tabcolsep}{2pt}
    \begin{tabular}{l c c c c c}
        \toprule
          Methods & \textbf{PSNR} & \textbf{SSIM} & \textbf{LPIPS} & \textbf{D-RMSE} & \textbf{Time} \\
         \midrule
          ZoeDepth & 5.47  & 0.25 &   0.56 &  11.73 & -- \\
          Metric3D  & 6.31  & 0.30 &  0.55 &  10.05 & -- \\
          NeRFacto   & 10.94  &  0.30 &  0.79 & -- & -- \\
          K-Planes  & 11.36  &  0.46 &   0.63 & -- & -- \\
          SplatFacto  & 11.61  &  0.49 &   0.66 &  -- & $\approx$  480s \\
          MVSplat  & 13.86 &  0.46 &  0.66 & 16.79 & \cellcolor{tabfirst}0.42ms \\
          PixelNeRF  & 14.50 & 0.55 & 0.65 & 19.24 & 1.86s\\
          SplatterImg.   & 17.79  & 0.58 &  0.57 & 11.05 & \cellcolor{tabthird}32ms  \\ 
          pixelSplat & \cellcolor{tabthird}18.03 & \cellcolor{tabthird}0.60 & \cellcolor{tabfirst}0.44 & \cellcolor{tabthird}7.26 &  \cellcolor{tabsecond}1.1ms \\ 
          6Img-to-3D  & \cellcolor{tabsecond}18.68 & \cellcolor{tabfirst}0.73 & \cellcolor{tabsecond}0.45 & \cellcolor{tabfirst}6.23 & 2.85s\\     
          \bottomrule
          sshELF (Ours) & \cellcolor{tabfirst}18.93 &  \cellcolor{tabsecond}0.65 & \cellcolor{tabthird}0.50  & \cellcolor{tabsecond}6.61 & 182ms\\
        \bottomrule
    \end{tabular}
    \caption{\textbf{SEED4D Results}. Runtime comparison of scene-to-novel view inference, presented in both seconds and milliseconds to account for the large variations in execution time across different methods.}
    \label{tab:results_seed4d}
\end{table}

\begin{figure}[!ht]
    \centering
    \includegraphics[width=1\linewidth]{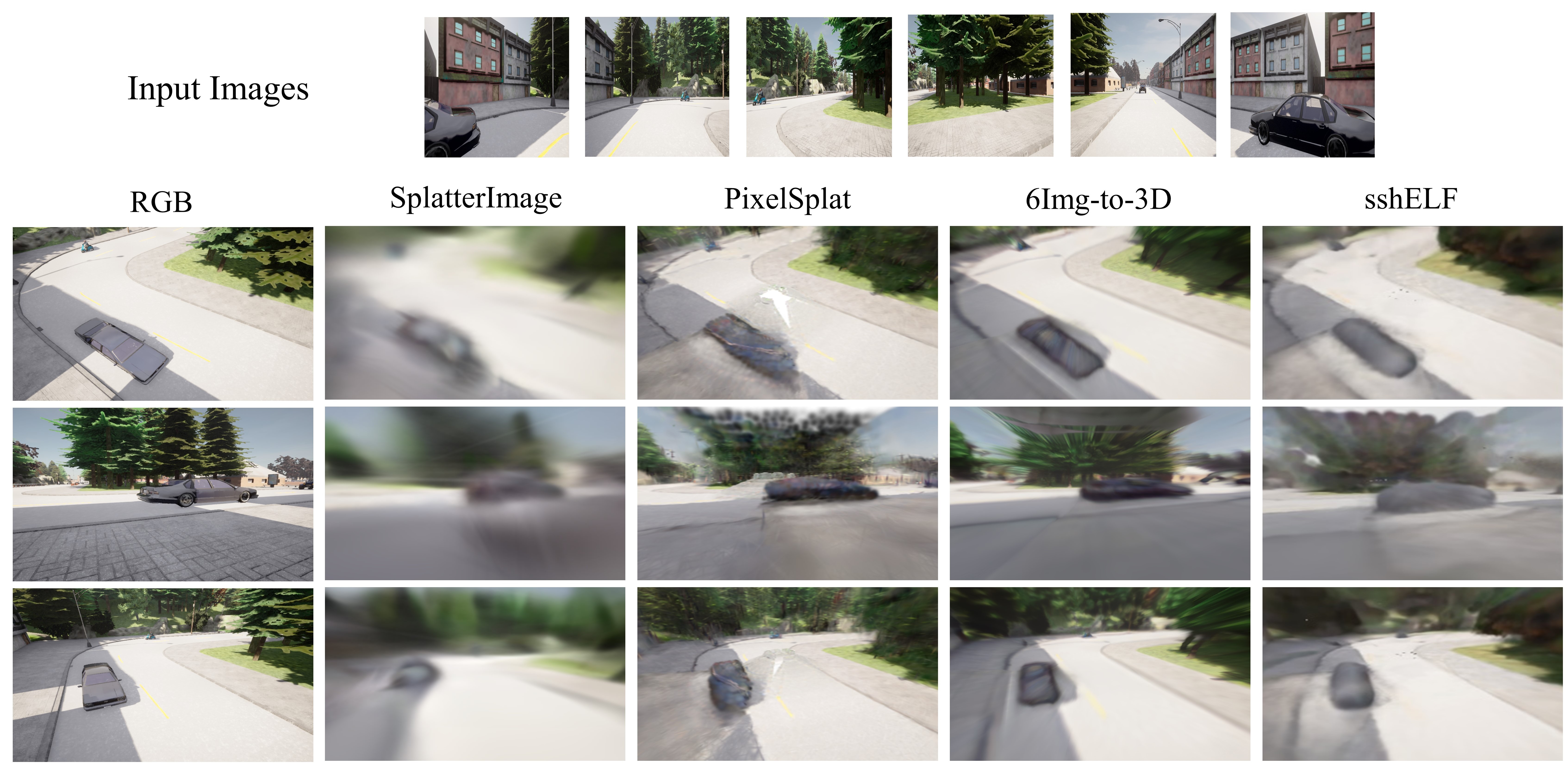}
    \caption{\textbf{Qualitative Novel View Synthesis Comparison on SEED4D Test Set}. Comparison of large-baseline novel view synthesis under sparse observation conditions. Six ego-centric input frames (top row) with limited overlap serve as reference views. We evaluate each method's ability to reconstruct exo-centric with a large offset to the input views.}
    \label{fig:seed4d_qualitiative}
\end{figure}

\noindent \textbf{NuScenes}. Quantitative multi-timestep results in Table~\ref{tab:results_nuscenes} demonstrate our method's superiority across visual and geometric metrics. The virtual view sampling in sshELF enables strategic allocation of scene representation capacity: by prioritizing reconstruction fidelity for distant viewpoints critical for wide-baseline tasks, our method inherently trades off minor quality reductions in near-field regions. In contrast methods like MVSplat~\cite{MVSplat} and PixelSplat~\cite{pixelsplat_Charatan_2024_CVPR}, which reprojecting input pixels onto local planes—a design that limits scalability to far-view synthesis. This approach fails to model occlusions or parallax effects at larger distances as seen in Figure \ref{fig:seed4d_qualitiative} for MVSplat. Compared to the baselines, our method more accurately represents color information and reconstructs occluded regions with greater fidelity.

\begin{table}[!ht] 
    \setlength{\tabcolsep}{2pt}
    \setlength{\tabcolsep}{2.5pt}
    \begin{tabular}{l l cccc }
        \toprule
          & Methods & \textbf{PSNR}$\uparrow$ & \textbf{SSIM}$\uparrow$ & \textbf{LPIPS}$\downarrow$ & \textbf{Chamfer}$\downarrow$ \\
         \midrule
       \multirow{3}{*}{\rotatebox{90}{\textbf{TD2}}} \rule{3pt}{0em} & MVSplat  & \cellcolor{tabthird}16.624 & \cellcolor{tabthird}0.436 & \cellcolor{tabsecond}0.553 & \cellcolor{tabsecond}646.11  \\
        & pixelSplat  & \cellcolor{tabsecond}18.031 & \cellcolor{tabsecond}0.459 & \cellcolor{tabfirst}0.495 & \cellcolor{tabthird}1.191M  \\
        \cmidrule{2-6} 
        &  sshELF (Ours) & \cellcolor{tabfirst}19.133 & \cellcolor{tabfirst}0.645 &  \cellcolor{tabthird}0.634 &  \cellcolor{tabfirst}51.67  \\
        \bottomrule
        \multirow{3}{*}{\rotatebox{90}{\textbf{TD3}}} & MVSplat  & \cellcolor{tabthird}16.551 & \cellcolor{tabsecond}0.448 & \cellcolor{tabsecond}0.575 & \cellcolor{tabsecond}2861.61 \\
        & pixelSplat  & \cellcolor{tabsecond}17.332 & \cellcolor{tabthird}0.426 & \cellcolor{tabfirst}0.532 &  \cellcolor{tabthird}0.144M  \\
        \cmidrule{2-6} 
        &  sshELF (Ours) & \cellcolor{tabfirst}18.174 & \cellcolor{tabfirst}0.635 & \cellcolor{tabthird}0.650 & \cellcolor{tabfirst}124.80 \\
        \bottomrule
        \multirow{3}{*}{\rotatebox{90}{\textbf{TD4}}} & MVSplat  & \cellcolor{tabthird}12.610 & \cellcolor{tabthird}0.380 & \cellcolor{tabthird}0.714 & \cellcolor{tabsecond}202.00 \\
        & pixelSplat  & \cellcolor{tabsecond}17.306 & \cellcolor{tabsecond}0.438 & \cellcolor{tabfirst}0.539 & \cellcolor{tabthird}0.163M \\
        \cmidrule{2-6} 
        &  sshELF (Ours) & \cellcolor{tabfirst}17.594 & \cellcolor{tabfirst}0.628 & \cellcolor{tabsecond}0.653 & \cellcolor{tabfirst}171.81 \\
        \bottomrule
    \end{tabular}
    \caption{\textbf{nuScenes Results}. Results are shown for temporal differences (TD) of 2, 3, and 4 in terms of PSNR, SSIM, LPIPS and Chamfer distance.}
     \label{tab:results_nuscenes}
\end{table}
\vspace{0.2cm}

\begin{figure}
    \centering
    \includegraphics[width=1\linewidth]{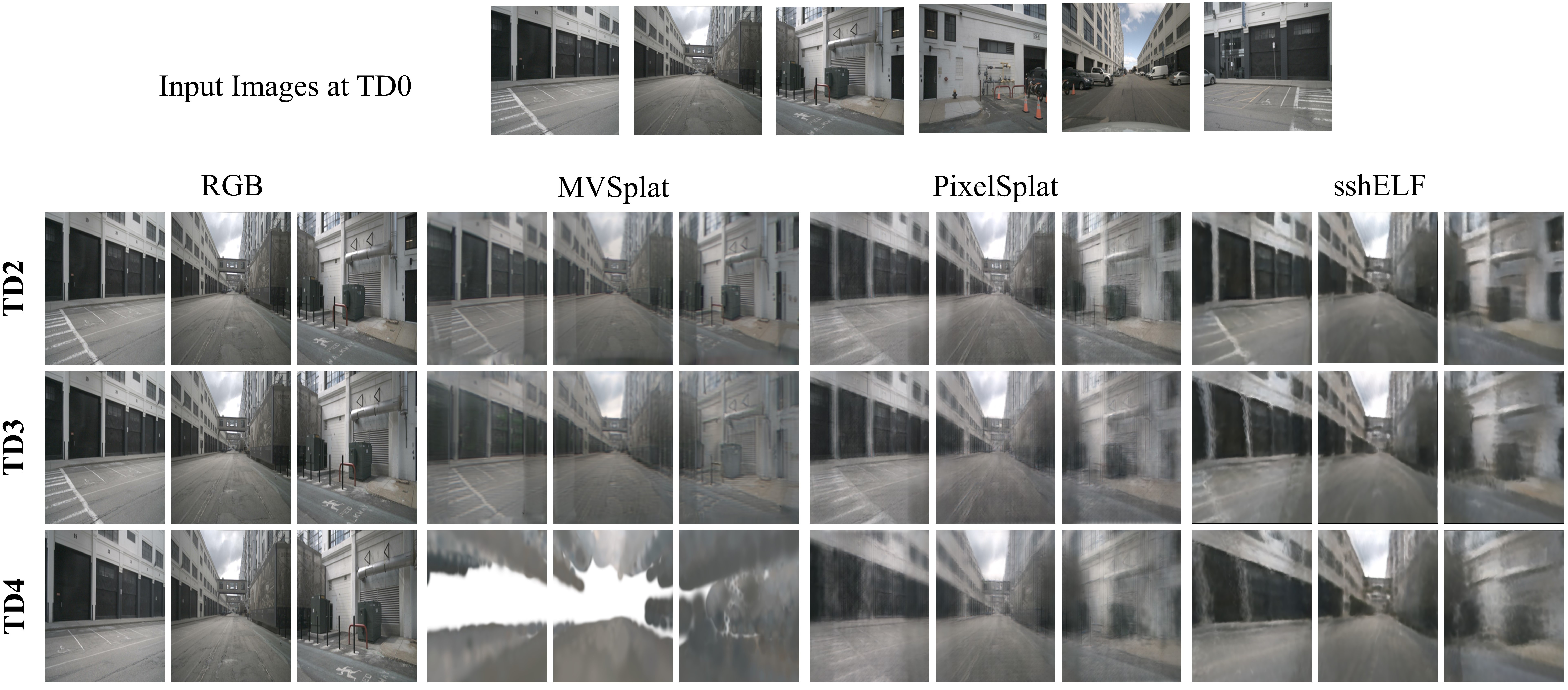}
    \caption{\textbf{Qualitative Novel View Synthesis Comparison on nuScenes Test Set}. Visualization of multi-view synthesis results using six reference views captured at t=0. We compare novel views reconstructed at temporal difference of TD=2, 3, and 4 (1s, 1.5s, and 2s, respectively).}
    \label{fig:nuScenes_qualitative}
\end{figure}


\subsection{Ablations and Analysis}\label{subsec:ablation}
The following questions are investigated:

\underline{Question 1}  Which ELF block architecture is best suited for reconstructing virtual views? \\
\underline{Question 2}: How impactful is the size of the reconstructed views for the overall model performance? \\
\underline{Question 3}: Does adding a Chamfer distance loss to the model loss improve the results? \\

\noindent \textbf{Backbone Design (Q1)}. We experiment with varying the number of cross-attention and self-attention blocks within each ELF block. Performance is evaluated using PSNR, SSIM, and LPIPS metrics on reconstructed virtual views. Each resulting backbone is trained for 60K steps. The results obtained on the SEED4D dataset are summarized in Table \ref{tab:results_backbone_performance_elf}.
\begin{table}[!ht]
    \centering
    \setlength{\tabcolsep}{5pt}
    \begin{tabular}{l c c c c c}
        \toprule
          \textbf{ELF} & \textbf{PSNR} $\uparrow$ & \textbf{SSIM}$\uparrow$ & \textbf{LPIPS}$\downarrow$ \\
          \midrule
          1 cc with 1 cs  & 16.3002  &  \cellcolor{tabthird}0.4049 &   0.6032  \\
          2 cc with 1 cs  & \cellcolor{tabfirst}16.5557  &  \cellcolor{tabfirst}0.4184 &  \cellcolor{tabthird}0.5990 \\
          3 cc with 1 cs  & \cellcolor{tabsecond}16.4709  &  0.4022 &   0.6044 \\
          3 cc with 3 cs  & 16.2464  &  \cellcolor{tabsecond}0.4053 &   \cellcolor{tabfirst}0.5980 \\
          4 cc with 1 cs  & \cellcolor{tabthird}16.3215  &  0.4017 &   \cellcolor{tabsecond}0.5982 \\
        \bottomrule
    \end{tabular}
    \caption{\textbf{Backbone Performance}. cc indicates the number of cross-attention blocks and cs the number of self-attention blocks.}
    \label{tab:results_backbone_performance_elf}
\end{table}

\noindent \textbf{Translator Design (Q2)}. We investigate the impact of input view resolution by varying the size of the views inputted into the translator. Additionally, we explore a different gradient propagation styles, as proposed in Depth Normalization Regularized Gaussians (DNG) \cite{li2024dngaussianoptimizingsparseview3d}. Table \ref{tab:translator_design}, presents the training results for each configuration after 60K steps on the SEED4D dataset, highlighting the influence of these design choices.
\begin{table} 
    \centering
    \setlength{\tabcolsep}{2pt}
    \begin{tabular}{l c c c c}
        \toprule
          Resolution & \textbf{PSNR} $\uparrow$ & \textbf{SSIM}$\uparrow$ & \textbf{LPIPS}$\downarrow$ \\
         \midrule
         64 w/o DNG  & 17.8219  &  \cellcolor{tabthird}0.5970 &  \cellcolor{tabthird}0.6089 \\
         64 with DNG  & \cellcolor{tabthird}17.9062  &  0.5947 &   0.6106 \\
         128 w/o DNG  & \cellcolor{tabsecond}17.9950  &  \cellcolor{tabsecond}0.6116 &  \cellcolor{tabsecond}0.6041 \\
         128 with DNG  & 15.2253  &  0.5437 &   0.6787 \\
         256 w/o DNG  &\cellcolor{tabfirst}18.1216  &  \cellcolor{tabfirst}0.6137 &  \cellcolor{tabfirst}0.6017 \\
         256 with DNG & 14.8780  &  0.5318 &   0.6802 \\
        \bottomrule
    \end{tabular}
    \caption{\textbf{Translator ablation}. Our analysis reveals that DNG underperforms, while reconstruction quality improves with higher image resolution.}
    \label{tab:translator_design}
\end{table}

\noindent \textbf{Chamfer Distance Loss (Q3)}. Since the nuScenes dataset includes LiDAR data, we compute the Chamfer distance and experiment with enforcing a Chamfer distance loss during training. While this loss improves alignment with the ground truth geometry, it results in a slight degradation of visual metrics. Detailed results are provided in Table \ref{tab:results_chamfer_distance}.
\vspace{0.3cm}

\begin{table}[!ht]
    \setlength{\tabcolsep}{2pt}
    \vspace{-0.2cm}
    \setlength{\tabcolsep}{2.5pt}
    \begin{tabular}{l l cccc }
        \toprule
          & Methods & \textbf{PSNR}$\uparrow$ & \textbf{SSIM}$\uparrow$ & \textbf{LPIPS}$\downarrow$ & \textbf{Chamfer}$\downarrow$ \\
         \midrule
       \multirow{2}{*}{\rotatebox{90}{\textbf{TD2}}} \rule{3pt}{0em} 
        & Ours w C & \cellcolor{tabsecond}18.980 & \cellcolor{tabsecond}0.645 & \cellcolor{tabsecond}0.641 & \cellcolor{tabfirst}10.89  \\
        & Ours w/o C &\cellcolor{tabfirst}19.133 & \cellcolor{tabfirst}0.645 & \cellcolor{tabfirst}0.634 &  \cellcolor{tabsecond}51.67  \\
        \bottomrule
        \multirow{2}{*}{\rotatebox{90}{\textbf{TD3}}}  & Ours w C & \cellcolor{tabsecond}18.133 & \cellcolor{tabsecond}0.636 & \cellcolor{tabsecond}0.654 & \cellcolor{tabfirst}13.30 \\
        & Ours w/o C & \cellcolor{tabfirst}18.174 & \cellcolor{tabfirst}0.635 & \cellcolor{tabfirst}0.650 & \cellcolor{tabsecond}124.80 \\
        \bottomrule
        \multirow{2}{*}{\rotatebox{90}{\textbf{TD4}}}  & Ours w C & \cellcolor{tabsecond}17.557 & \cellcolor{tabsecond}0.628 & \cellcolor{tabsecond}0.658 & \cellcolor{tabfirst}13.51 \\
        & Ours w/o C & \cellcolor{tabfirst}17.594 & \cellcolor{tabfirst}0.628 & \cellcolor{tabfirst}0.653 & \cellcolor{tabsecond}171.81 \\
        \bottomrule
    \end{tabular}
    \caption{\textbf{Chamfer Distance Loss}. Results are shown for temporal differences (TD) of 2, 3, and 4 in terms of PSNR, SSIM, LPIPS, and Chamfer distance.}
    \label{tab:results_chamfer_distance}
\end{table}

\section{Conclusion}
\label{sec:dicussion}

This paper introduces sshELF, a fast 3D Gaussian-based framework for reconstructing unbounded driving scenes from sparse, outward-facing views. Our method overcomes the critical challenge of reconstructing unobserved regions, such as distant object occlusions and ego-occlusions, that existing approaches fail to resolve. Our key innovation lies in decoupling information extrapolation from primitive decoding, enabling cross-scene transfer of structural patterns while maintaining a modular, real-time capable pipeline.

Experiments on challenging synthetic and real-world datasets demonstrate that sshELF achieves competitive novel view synthesis results, even for heavily occluded regions. A current limitation is sensitivity to dynamic objects when aggregating multi-timestep data, which can introduce transient artifacts.
Future work will focus on (1) temporal filtering to mask dynamic objects during training, and (2) exploring the downstream performance of the obtained latent features. \\

\noindent \textbf{Impact Statement}. This paper presents work whose goal is to advance the field of Machine Learning. There are many potential societal consequences of our work, none of which we feel must be specifically highlighted here.

\noindent \paragraph{Acknowledgements}. The research leading to these results is partially funded by the German Federal Ministry for Economic Affairs and Climate Action within the project “NXT GEN AI METHODS".


\bibliography{example_paper}

\begin{thebibliography}{63}
\providecommand{\natexlab}[1]{#1}
\providecommand{\url}[1]{\texttt{#1}}
\expandafter\ifx\csname urlstyle\endcsname\relax
  \providecommand{\doi}[1]{doi: #1}\else
  \providecommand{\doi}{doi: \begingroup \urlstyle{rm}\Url}\fi

\bibitem[Anciukevi\v{c}ius et~al.(2023)Anciukevi\v{c}ius, Xu, Fisher, Henderson, Bilen, Mitra, and Guerrero]{RenderDiffusion_Anciukevicius_2023_CVPR}
Anciukevi\v{c}ius, T., Xu, Z., Fisher, M., Henderson, P., Bilen, H., Mitra, N.~J., and Guerrero, P.
\newblock Renderdiffusion: Image diffusion for 3d reconstruction, inpainting and generation.
\newblock In \emph{Proceedings of the IEEE/CVF Conference on Computer Vision and Pattern Recognition (CVPR)}, pp.\  12608--12618, June 2023.

\bibitem[Behley et~al.(2019)Behley, Garbade, Milioto, Quenzel, Behnke, Stachniss, and Gall]{behley_semantickitti_2019}
Behley, J., Garbade, M., Milioto, A., Quenzel, J., Behnke, S., Stachniss, C., and Gall, J.
\newblock Semantickitti: A dataset for semantic scene understanding of lidar sequences.
\newblock In \emph{2019 IEEE/CVF International Conference on Computer Vision (ICCV)}, pp.\  9296--9306, 2019.
\newblock \doi{10.1109/ICCV.2019.00939}.

\bibitem[Bhat et~al.(2023)Bhat, Birkl, Wofk, Wonka, and Müller]{bhat2023zoedepth}
Bhat, S.~F., Birkl, R., Wofk, D., Wonka, P., and Müller, M.
\newblock Zoedepth: Zero-shot transfer by combining relative and metric depth.
\newblock \emph{CoRR}, abs/2302.12288, 2023.

\bibitem[Caesar et~al.(2020)Caesar, Bankiti, Lang, Vora, Liong, Xu, Krishnan, Pan, Baldan, and Beijbom]{caesar_nuscenes_2020}
Caesar, H., Bankiti, V., Lang, A.~H., Vora, S., Liong, V.~E., Xu, Q., Krishnan, A., Pan, Y., Baldan, G., and Beijbom, O.
\newblock nuscenes: A multimodal dataset for autonomous driving.
\newblock In \emph{CVPR}, 2020.

\bibitem[Charatan et~al.(2024)Charatan, Li, Tagliasacchi, and Sitzmann]{pixelsplat_Charatan_2024_CVPR}
Charatan, D., Li, S.~L., Tagliasacchi, A., and Sitzmann, V.
\newblock pixelsplat: 3d gaussian splats from image pairs for scalable generalizable 3d reconstruction.
\newblock In \emph{Proceedings of the IEEE/CVF Conference on Computer Vision and Pattern Recognition (CVPR)}, pp.\  19457--19467, June 2024.

\bibitem[Chen et~al.(2021)Chen, Xu, Zhao, Zhang, Xiang, Yu, and Su]{mvsnerf}
Chen, A., Xu, Z., Zhao, F., Zhang, X., Xiang, F., Yu, J., and Su, H.
\newblock Mvsnerf: Fast generalizable radiance field reconstruction from multi-view stereo.
\newblock In \emph{Proceedings of the IEEE/CVF International Conference on Computer Vision}, pp.\  14124--14133, 2021.

\bibitem[Chen et~al.(2024{\natexlab{a}})Chen, Xu, Esposito, Tang, and Geiger]{LaRa}
Chen, A., Xu, H., Esposito, S., Tang, S., and Geiger, A.
\newblock Lara: Efficient large-baseline radiance fields.
\newblock In \emph{European Conference on Computer Vision (ECCV)}, 2024{\natexlab{a}}.

\bibitem[Chen et~al.(2023)Chen, Gu, Jiang, Zhu, and Zhang]{chen2023periodic}
Chen, Y., Gu, C., Jiang, J., Zhu, X., and Zhang, L.
\newblock Periodic vibration gaussian: Dynamic urban scene reconstruction and real-time rendering.
\newblock \emph{arXiv:2311.18561}, 2023.

\bibitem[Chen et~al.(2025)Chen, Xu, Zheng, Zhuang, Pollefeys, Geiger, Cham, and Cai]{MVSplat}
Chen, Y., Xu, H., Zheng, C., Zhuang, B., Pollefeys, M., Geiger, A., Cham, T.-J., and Cai, J.
\newblock Mvsplat: Efficient 3d gaussian splatting from sparse multi-view images.
\newblock In Leonardis, A., Ricci, E., Roth, S., Russakovsky, O., Sattler, T., and Varol, G. (eds.), \emph{Computer Vision -- ECCV 2024}, pp.\  370--386, Cham, 2025. Springer Nature Switzerland.

\bibitem[Chen et~al.(2024{\natexlab{b}})Chen, Yang, Huang, Lutio, Esturo, Ivanovic, Litany, Gojcic, Fidler, Pavone, Song, and Wang]{chen2024omnire}
Chen, Z., Yang, J., Huang, J., Lutio, R.~d., Esturo, J.~M., Ivanovic, B., Litany, O., Gojcic, Z., Fidler, S., Pavone, M., Song, L., and Wang, Y.
\newblock Omnire: Omni urban scene reconstruction.
\newblock \emph{arXiv preprint arXiv:2408.16760}, 2024{\natexlab{b}}.

\bibitem[Deitke et~al.(2023)Deitke, Schwenk, Salvador, Weihs, Michel, VanderBilt, Schmidt, Ehsanit, Kembhavi, and Farhadi]{deitke2022objaverseuniverseannotated3d}
Deitke, M., Schwenk, D., Salvador, J., Weihs, L., Michel, O., VanderBilt, E., Schmidt, L., Ehsanit, K., Kembhavi, A., and Farhadi, A.
\newblock Objaverse: A universe of annotated 3d objects.
\newblock In \emph{2023 IEEE/CVF Conference on Computer Vision and Pattern Recognition (CVPR)}, pp.\  13142--13153, Los Alamitos, CA, USA, jun 2023. IEEE Computer Society.
\newblock \doi{10.1109/CVPR52729.2023.01263}.

\bibitem[Dosovitskiy et~al.(2021)Dosovitskiy, Beyer, Kolesnikov, Weissenborn, Zhai, Unterthiner, Dehghani, Minderer, Heigold, Gelly, Uszkoreit, and Houlsby]{dosovitskiy2020vit}
Dosovitskiy, A., Beyer, L., Kolesnikov, A., Weissenborn, D., Zhai, X., Unterthiner, T., Dehghani, M., Minderer, M., Heigold, G., Gelly, S., Uszkoreit, J., and Houlsby, N.
\newblock An image is worth 16x16 words: Transformers for image recognition at scale.
\newblock \emph{ICLR}, 2021.

\bibitem[Fischer et~al.(2024)Fischer, Porzi, Rota~Bul\`{o}, Pollefeys, and Kontschieder]{ml_nsg_2024_CVPR}
Fischer, T., Porzi, L., Rota~Bul\`{o}, S., Pollefeys, M., and Kontschieder, P.
\newblock Multi-level neural scene graphs for dynamic urban environments.
\newblock In \emph{Proceedings of the IEEE/CVF Conference on Computer Vision and Pattern Recognition (CVPR)}, 2024.

\bibitem[Gieruc et~al.(2024)Gieruc, Kästingschäfer, Bernhard, and Salzmann]{gieruc20246imgto3d}
Gieruc, T., Kästingschäfer, M., Bernhard, S., and Salzmann, M.
\newblock 6img-to-3d: Few-image large-scale outdoor driving scene reconstruction.
\newblock \emph{arXiv preprint}, arXiv:2404.12378, 2024.

\bibitem[Guizilini et~al.(2022)Guizilini, Vasiljevic, Ambrus, Shakhnarovich, and Gaidon]{surround_monodepth2022}
Guizilini, V., Vasiljevic, I., Ambrus, R., Shakhnarovich, G., and Gaidon, A.
\newblock Full surround monodepth from multiple cameras.
\newblock \emph{IEEE Robotics and Automation Letters}, 7\penalty0 (2):\penalty0 5397--5404, 2022.
\newblock \doi{10.1109/LRA.2022.3150884}.

\bibitem[Guo et~al.(2023)Guo, Deng, Li, Bai, Shi, Wang, Ding, Wang, and Li]{guo2023streetsurf}
Guo, J., Deng, N., Li, X., Bai, Y., Shi, B., Wang, C., Ding, C., Wang, D., and Li, Y.
\newblock Streetsurf: Extending multi-view implicit surface reconstruction to street views.
\newblock \emph{arXiv preprint arXiv:2306.04988}, 2023.

\bibitem[Hwang et~al.(2024)Hwang, Kim, Kang, Kang, and Choo]{VEGS}
Hwang, S., Kim, M., Kang, T., Kang, J., and Choo, J.
\newblock {VEGS:} view extrapolation of urban scenes in 3d gaussian splatting using learned priors.
\newblock \emph{CoRR}, abs/2407.02945, 2024.
\newblock \doi{10.48550/ARXIV.2407.02945}.

\bibitem[Irshad et~al.(2023)Irshad, Zakharov, Liu, Guizilini, Kollar, Gaidon, Kira, and Ambrus]{irshad2023neo360}
Irshad, M.~Z., Zakharov, S., Liu, K., Guizilini, V., Kollar, T., Gaidon, A., Kira, Z., and Ambrus, R.
\newblock Neo 360: Neural fields for sparse view synthesis of outdoor scenes.
\newblock \emph{Interntaional Conference on Computer Vision (ICCV)}, 2023.

\bibitem[Johnson et~al.(2020)Johnson, Ravi, Reizenstein, Novotny, Tulsiani, Lassner, and Branson]{ravi2020pytorch3d}
Johnson, J., Ravi, N., Reizenstein, J., Novotny, D., Tulsiani, S., Lassner, C., and Branson, S.
\newblock Accelerating 3d deep learning with pytorch3d.
\newblock In \emph{SIGGRAPH Asia 2020 Courses}, SA '20, New York, NY, USA, 2020. Association for Computing Machinery.
\newblock ISBN 9781450381123.
\newblock \doi{10.1145/3415263.3419160}.

\bibitem[Ke et~al.(2024)Ke, Obukhov, Huang, Metzger, Daudt, and Schindler]{marigold_ke2023repurposing}
Ke, B., Obukhov, A., Huang, S., Metzger, N., Daudt, R.~C., and Schindler, K.
\newblock Repurposing diffusion-based image generators for monocular depth estimation.
\newblock In \emph{Proceedings of the IEEE/CVF Conference on Computer Vision and Pattern Recognition (CVPR)}, 2024.

\bibitem[Kerbl et~al.(2023)Kerbl, Kopanas, Leimk{\"u}hler, and Drettakis]{kerbl3Dgaussians}
Kerbl, B., Kopanas, G., Leimk{\"u}hler, T., and Drettakis, G.
\newblock 3d gaussian splatting for real-time radiance field rendering.
\newblock \emph{ACM Transactions on Graphics}, 42\penalty0 (4), July 2023.

\bibitem[Khan et~al.(2024)Khan, Fazlali, Sharma, Cao, Bai, Ren, and Liu]{khan2024autosplatconstrainedgaussiansplatting}
Khan, M., Fazlali, H., Sharma, D., Cao, T., Bai, D., Ren, Y., and Liu, B.
\newblock Autosplat: Constrained gaussian splatting for autonomous driving scene reconstruction.
\newblock \emph{arXiv preprint}, arXiv:2407.02598, 2024.

\bibitem[Kingma \& Ba(2015)Kingma and Ba]{Kingma_Adam}
Kingma, D.~P. and Ba, J.
\newblock Adam: {A} method for stochastic optimization.
\newblock In Bengio, Y. and LeCun, Y. (eds.), \emph{3rd International Conference on Learning Representations, {ICLR} 2015, San Diego, CA, USA, May 7-9, 2015, Conference Track Proceedings}, 2015.

\bibitem[Kästingschäfer et~al.(2025)Kästingschäfer, Gieruc, Bernhard, Campbell, Insafutdinov, Najafli, and Brox]{SEED4D}
Kästingschäfer, M., Gieruc, T., Bernhard, S., Campbell, D., Insafutdinov, E., Najafli, E., and Brox, T.
\newblock Seed4d: A synthetic ego--exo dynamic 4d data generator, driving dataset and benchmark.
\newblock \emph{arXiv preprint}, arXiv:2412.00730, 2025.
\newblock URL \url{https://arxiv.org/abs/2412.00730}.

\bibitem[Lefaudeux et~al.(2022)Lefaudeux, Massa, Liskovich, Xiong, Caggiano, Naren, Xu, Hu, Tintore, Zhang, Labatut, Haziza, Wehrstedt, Reizenstein, and Sizov]{xFormers2022}
Lefaudeux, B., Massa, F., Liskovich, D., Xiong, W., Caggiano, V., Naren, S., Xu, M., Hu, J., Tintore, M., Zhang, S., Labatut, P., Haziza, D., Wehrstedt, L., Reizenstein, J., and Sizov, G.
\newblock xformers: A modular and hackable transformer modelling library.
\newblock \url{https://github.com/facebookresearch/xformers}, 2022.

\bibitem[Li et~al.(2024{\natexlab{a}})Li, Li, Zhang, Wu, Shi, Zhao, Feng, Ding, Wang, and Han]{li2024_VDG}
Li, H., Li, J., Zhang, D., Wu, C., Shi, J., Zhao, C., Feng, H., Ding, E., Wang, J., and Han, J.
\newblock Vdg: Vision-only dynamic gaussian for driving simulation.
\newblock \emph{arXiv preprint}, 2024{\natexlab{a}}.

\bibitem[Li et~al.(2024{\natexlab{b}})Li, Zhang, Bai, Zheng, Ning, Zhou, and Gu]{li2024dngaussianoptimizingsparseview3d}
Li, J., Zhang, J., Bai, X., Zheng, J., Ning, X., Zhou, J., and Gu, L.
\newblock Dngaussian: Optimizing sparse-view 3d gaussian radiance fields with global-local depth normalization.
\newblock \emph{arXiv preprint}, arXiv:2403.06912, 2024{\natexlab{b}}.
\newblock URL \url{https://arxiv.org/abs/2403.06912}.

\bibitem[Li et~al.(2023)Li, Li, and Zhu]{li2022read}
Li, Z., Li, L., and Zhu, J.
\newblock Read: Large-scale neural scene rendering for autonomous driving.
\newblock In \emph{AAAI}, 2023.

\bibitem[Mildenhall et~al.(2020)Mildenhall, Srinivasan, Tancik, Barron, Ramamoorthi, and Ng]{mildenhall2020nerf}
Mildenhall, B., Srinivasan, P.~P., Tancik, M., Barron, J.~T., Ramamoorthi, R., and Ng, R.
\newblock Nerf: Representing scenes as neural radiance fields for view synthesis.
\newblock In \emph{ECCV}, 2020.

\bibitem[Oquab et~al.(2023)Oquab, Darcet, Moutakanni, Vo, Szafraniec, Khalidov, Fernandez, Haziza, Massa, El-Nouby, Howes, Huang, Xu, Sharma, Li, Galuba, Rabbat, Assran, Ballas, Synnaeve, Misra, Jegou, Mairal, Labatut, Joulin, and Bojanowski]{oquab2023dinov2}
Oquab, M., Darcet, T., Moutakanni, T., Vo, H.~V., Szafraniec, M., Khalidov, V., Fernandez, P., Haziza, D., Massa, F., El-Nouby, A., Howes, R., Huang, P.-Y., Xu, H., Sharma, V., Li, S.-W., Galuba, W., Rabbat, M., Assran, M., Ballas, N., Synnaeve, G., Misra, I., Jegou, H., Mairal, J., Labatut, P., Joulin, A., and Bojanowski, P.
\newblock Dinov2: Learning robust visual features without supervision.
\newblock \emph{arXiv preprint}, arXiv:2304.07193, 2023.

\bibitem[Ost et~al.(2021)Ost, Mannan, Thuerey, Knodt, and Heide]{NSG_2021_CVPR}
Ost, J., Mannan, F., Thuerey, N., Knodt, J., and Heide, F.
\newblock Neural scene graphs for dynamic scenes.
\newblock In \emph{Proceedings of the IEEE/CVF Conference on Computer Vision and Pattern Recognition (CVPR)}, pp.\  2856--2865, June 2021.

\bibitem[Paszke et~al.(2019)Paszke, Gross, Massa, Lerer, Bradbury, Chanan, Killeen, Lin, Gimelshein, Antiga, Desmaison, Kopf, Yang, DeVito, Raison, Tejani, Chilamkurthy, Steiner, Fang, Bai, and Chintala]{PyTorch_NeurIPS2019}
Paszke, A., Gross, S., Massa, F., Lerer, A., Bradbury, J., Chanan, G., Killeen, T., Lin, Z., Gimelshein, N., Antiga, L., Desmaison, A., Kopf, A., Yang, E., DeVito, Z., Raison, M., Tejani, A., Chilamkurthy, S., Steiner, B., Fang, L., Bai, J., and Chintala, S.
\newblock Pytorch: An imperative style, high-performance deep learning library.
\newblock In \emph{Advances in Neural Information Processing Systems 32}, pp.\  8024--8035. Curran Associates, Inc., 2019.

\bibitem[Ranftl et~al.(2021)Ranftl, Bochkovskiy, and Koltun]{DPT_Ranftl_2021_ICCV}
Ranftl, R., Bochkovskiy, A., and Koltun, V.
\newblock Vision transformers for dense prediction.
\newblock In \emph{Proceedings of the IEEE/CVF International Conference on Computer Vision (ICCV)}, pp.\  12179--12188, October 2021.

\bibitem[Reizenstein et~al.(2021)Reizenstein, Shapovalov, Henzler, Sbordone, Labatut, and Novotny]{reizenstein2021commonobjects3dlargescale}
Reizenstein, J., Shapovalov, R., Henzler, P., Sbordone, L., Labatut, P., and Novotny, D.
\newblock Common objects in 3d: Large-scale learning and evaluation of real-life 3d category reconstruction.
\newblock In \emph{2021 IEEE/CVF International Conference on Computer Vision (ICCV)}, pp.\  10881--10891, Los Alamitos, CA, USA, oct 2021. IEEE Computer Society.
\newblock \doi{10.1109/ICCV48922.2021.01072}.

\bibitem[Ronneberger et~al.(2015)Ronneberger, Fischer, and Brox]{unet_2015}
Ronneberger, O., Fischer, P., and Brox, T.
\newblock U-net: Convolutional networks for biomedical image segmentation.
\newblock \emph{CoRR}, abs/1505.04597, 2015.

\bibitem[Song et~al.(2021)Song, Sohl-Dickstein, Kingma, Kumar, Ermon, and Poole]{song2021scorebased}
Song, Y., Sohl-Dickstein, J., Kingma, D.~P., Kumar, A., Ermon, S., and Poole, B.
\newblock Score-based generative modeling through stochastic differential equations.
\newblock In \emph{International Conference on Learning Representations}, 2021.

\bibitem[Sun et~al.(2020)Sun, Kretzschmar, Dotiwalla, Chouard, Patnaik, Tsui, Guo, Zhou, Chai, Caine, Vasudevan, Han, Ngiam, Zhao, Timofeev, Ettinger, Krivokon, Gao, Joshi, Zhang, Shlens, Chen, and Anguelov]{Sun_2020_CVPR}
Sun, P., Kretzschmar, H., Dotiwalla, X., Chouard, A., Patnaik, V., Tsui, P., Guo, J., Zhou, Y., Chai, Y., Caine, B., Vasudevan, V., Han, W., Ngiam, J., Zhao, H., Timofeev, A., Ettinger, S., Krivokon, M., Gao, A., Joshi, A., Zhang, Y., Shlens, J., Chen, Z., and Anguelov, D.
\newblock Scalability in perception for autonomous driving: Waymo open dataset.
\newblock In \emph{Proceedings of the IEEE/CVF Conference on Computer Vision and Pattern Recognition (CVPR)}, June 2020.

\bibitem[Szymanowicz et~al.(2023)Szymanowicz, Rupprecht, and Vedaldi]{szymanowicz2023viewset_diffusion}
Szymanowicz, S., Rupprecht, C., and Vedaldi, A.
\newblock Viewset diffusion: (0-)image-conditioned 3d generative models from 2d data.
\newblock \emph{International Conference on Computer Vision}, 2023.

\bibitem[Szymanowicz et~al.(2024{\natexlab{a}})Szymanowicz, Insafutdinov, Zheng, Campbell, Henriques, Rupprecht, and Vedaldi]{szymanowicz2024flash3d}
Szymanowicz, S., Insafutdinov, E., Zheng, C., Campbell, D., Henriques, J., Rupprecht, C., and Vedaldi, A.
\newblock Flash3d: Feed-forward generalisable 3d scene reconstruction from a single image.
\newblock \emph{arxiv}, 2024{\natexlab{a}}.

\bibitem[Szymanowicz et~al.(2024{\natexlab{b}})Szymanowicz, Rupprecht, and Vedaldi]{szymanowicz24splatter}
Szymanowicz, S., Rupprecht, C., and Vedaldi, A.
\newblock Splatter image: Ultra-fast single-view 3d reconstruction.
\newblock In \emph{The IEEE/CVF Conference on Computer Vision and Pattern Recognition (CVPR)}, 2024{\natexlab{b}}.

\bibitem[Tian et~al.(2024)Tian, Tan, Xie, and Ma]{tian2024drivingforwardfeedforward3dgaussian}
Tian, Q., Tan, X., Xie, Y., and Ma, L.
\newblock Drivingforward: Feed-forward 3d gaussian splatting for driving scene reconstruction from flexible surround-view input.
\newblock \emph{arXiv preprint}, arXiv:2409.12753, 2024.

\bibitem[Tonderski et~al.(2024)Tonderski, Lindstr\"om, Hess, Ljungbergh, Svensson, and Petersson]{NeuRAD_Tonderski_2024_CVPR}
Tonderski, A., Lindstr\"om, C., Hess, G., Ljungbergh, W., Svensson, L., and Petersson, C.
\newblock Neurad: Neural rendering for autonomous driving.
\newblock In \emph{Proceedings of the IEEE/CVF Conference on Computer Vision and Pattern Recognition (CVPR)}, pp.\  14895--14904, June 2024.

\bibitem[Turki et~al.(2023)Turki, Y, and Ferroni]{turki2023suds}
Turki, H., Y, Z., and Ferroni, Francesco~Ramanan, D.
\newblock Suds: Scalable urban dynascenes.
\newblock In \emph{Computer Vision Pattern Recognition (CVPR)}, 2023.

\bibitem[Wang et~al.(2024)Wang, Kim, Yang, Yu, Ivanovic, Waslander, Wang, Fidler, Pavone, and Karkus]{wang2024distillnerf}
Wang, L., Kim, S.~W., Yang, J., Yu, C., Ivanovic, B., Waslander, S.~L., Wang, Y., Fidler, S., Pavone, M., and Karkus, P.
\newblock Distillnerf: Perceiving 3d scenes from single-glance images by distilling neural fields and foundation model features.
\newblock In \emph{Conference on Neural Information Processing Systems}, 2024.

\bibitem[Wang et~al.(2021)Wang, Wang, Genova, Srinivasan, Zhou, Barron, Martin-Brualla, Snavely, and Funkhouser]{wang2021ibrnet}
Wang, Q., Wang, Z., Genova, K., Srinivasan, P., Zhou, H., Barron, J.~T., Martin-Brualla, R., Snavely, N., and Funkhouser, T.
\newblock Ibrnet: Learning multi-view image-based rendering.
\newblock In \emph{CVPR}, 2021.

\bibitem[Wang et~al.(2004)Wang, Bovik, Sheikh, and Simoncelli]{ssim}
Wang, Z., Bovik, A.~C., Sheikh, H.~R., and Simoncelli, E.~P.
\newblock Image quality assessment: from error visibility to structural similarity.
\newblock \emph{IEEE Transactions on Image Processing}, 13\penalty0 (4):\penalty0 600--612, 2004.

\bibitem[Wewer et~al.(2024)Wewer, Raj, Ilg, Schiele, and Lenssen]{wewer2024latentsplat}
Wewer, C., Raj, K., Ilg, E., Schiele, B., and Lenssen, J.~E.
\newblock {{\textbraceleft}latentSplat{\textbraceright}: {\textbraceleft}A{\textbraceright}utoencoding Variational {\textbraceleft}G{\textbraceright}aussians for Fast Generalizable {\textbraceleft}3D{\textbraceright} Reconstruction}.
\newblock In \emph{{Computer Vision -- ECCV 2024}}, {Lecture Notes in Computer Science}, Milano, Italy, 2024. Springer.
\newblock 18th European Conference on Computer Vision.

\bibitem[Wu et~al.(2024)Wu, Mildenhall, Henzler, Park, Gao, Watson, Srinivasan, Verbin, Barron, Poole, and Ho?y?ski]{ReconFusion_Wu_2024_CVPR}
Wu, R., Mildenhall, B., Henzler, P., Park, K., Gao, R., Watson, D., Srinivasan, P.~P., Verbin, D., Barron, J.~T., Poole, B., and Ho?y?ski, A.
\newblock Reconfusion: 3d reconstruction with diffusion priors.
\newblock In \emph{Proceedings of the IEEE/CVF Conference on Computer Vision and Pattern Recognition (CVPR)}, pp.\  21551--21561, June 2024.

\bibitem[Wu et~al.(2023)Wu, Liu, Luo, Zhong, Chen, Xiao, Hou, Lou, Chen, Yang, Huang, Ye, Yan, Shi, Liao, and Zhao]{wu2023mars}
Wu, Z., Liu, T., Luo, L., Zhong, Z., Chen, J., Xiao, H., Hou, C., Lou, H., Chen, Y., Yang, R., Huang, Y., Ye, X., Yan, Z., Shi, Y., Liao, Y., and Zhao, H.
\newblock Mars: An instance-aware, modular and realistic simulator for autonomous driving.
\newblock \emph{CICAI}, 2023.

\bibitem[Yan et~al.(2024)Yan, Lin, Zhou, Wang, Sun, Zhan, Lang, Zhou, and Peng]{yan2024street}
Yan, Y., Lin, H., Zhou, C., Wang, W., Sun, H., Zhan, K., Lang, X., Zhou, X., and Peng, S.
\newblock Street gaussians: Modeling dynamic urban scenes with gaussian splatting.
\newblock In \emph{ECCV}, 2024.

\bibitem[Yang et~al.(2024{\natexlab{a}})Yang, Li, Fang, Liang, Xie, Zhang, Shen, and Tian]{yang2024gaussianobject}
Yang, C., Li, S., Fang, J., Liang, R., Xie, L., Zhang, X., Shen, W., and Tian, Q.
\newblock Gaussianobject: High-quality 3d object reconstruction from four views with gaussian splatting.
\newblock \emph{ACM Transactions on Graphics}, 2024{\natexlab{a}}.

\bibitem[Yang et~al.(2024{\natexlab{b}})Yang, Ivanovic, Litany, Weng, Kim, Li, Che, Xu, Fidler, Pavone, and Wang]{yang2023emernerf}
Yang, J., Ivanovic, B., Litany, O., Weng, X., Kim, S.~W., Li, B., Che, T., Xu, D., Fidler, S., Pavone, M., and Wang, Y.
\newblock Emernerf: Emergent spatial-temporal scene decomposition via self-supervision.
\newblock In \emph{International Conference on Learning Representations}, 2024{\natexlab{b}}.

\bibitem[Yang et~al.(2024{\natexlab{c}})Yang, Kang, Huang, Xu, Feng, and Zhao]{depth_anything_v2}
Yang, L., Kang, B., Huang, Z., Xu, X., Feng, J., and Zhao, H.
\newblock Depth anything: Unleashing the power of large-scale unlabeled data.
\newblock In \emph{{IEEE/CVF} Conference on Computer Vision and Pattern Recognition, {CVPR} 2024, Seattle, WA, USA, June 16-22, 2024}, pp.\  10371--10381. {IEEE}, 2024{\natexlab{c}}.
\newblock \doi{10.1109/CVPR52733.2024.00987}.

\bibitem[Yang et~al.(2024{\natexlab{d}})Yang, Wang, Li, Tian, Sirasao, and Yang]{yang2024scaleawaresurroundmonodepthtransformers}
Yang, Y., Wang, X., Li, D., Tian, L., Sirasao, A., and Yang, X.
\newblock Towards scale-aware full surround monodepth with transformers.
\newblock \emph{arXiv preprint}, arXiv:2407.10406, 2024{\natexlab{d}}.

\bibitem[Yin et~al.(2023)Yin, Zhang, Chen, Cai, Yu, Wang, Chen, and Shen]{yin2023metric3d}
Yin, W., Zhang, C., Chen, H., Cai, Z., Yu, G., Wang, K., Chen, X., and Shen, C.
\newblock Metric3d: Towards zero-shot metric 3d prediction from a single image.
\newblock In \emph{Proceedings of the IEEE/CVF International Conference on Computer Vision}, pp.\  9043--9053, 2023.

\bibitem[Yinghao et~al.(2024)Yinghao, Zifan, Wang, Hansheng, Ceyuan, Sida, Yujun, and Gordon]{xu2024grm}
Yinghao, X., Zifan, S., Wang, Y., Hansheng, C., Ceyuan, Y., Sida, P., Yujun, S., and Gordon, W.
\newblock Grm: Large gaussian reconstruction model for efficient 3d reconstruction and generation.
\newblock \emph{arXiv preprint}, arXiv:2403.14621, 2024.

\bibitem[Yu et~al.(2021)Yu, Ye, Tancik, and Kanazawa]{yu2020pixelnerf}
Yu, A., Ye, V., Tancik, M., and Kanazawa, A.
\newblock {pixelNeRF}: Neural radiance fields from one or few images.
\newblock In \emph{CVPR}, 2021.

\bibitem[Yu et~al.(2024)Yu, Wang, Yang, Wang, Xie, Cai, Cao, Ji, and Sun]{Yu2024SGDSV}
Yu, Z., Wang, H., Yang, J., Wang, H., Xie, Z., Cai, Y., Cao, J., Ji, Z., and Sun, M.
\newblock Sgd: Street view synthesis with gaussian splatting and diffusion prior.
\newblock \emph{ArXiv}, abs/2403.20079, 2024.

\bibitem[Zhang et~al.(2018)Zhang, Isola, Efros, Shechtman, and Wang]{zhang2018unreasonable}
Zhang, R., Isola, P., Efros, A.~A., Shechtman, E., and Wang, O.
\newblock The unreasonable effectiveness of deep features as a perceptual metric.
\newblock \emph{2018 IEEE/CVF Conference on Computer Vision and Pattern Recognition}, pp.\  586--595, 2018.

\bibitem[Zhou et~al.(2024{\natexlab{a}})Zhou, Shao, Xu, Bai, Qiu, Liu, Wang, Geiger, and Liao]{Hugs_Zhou_2024_CVPR}
Zhou, H., Shao, J., Xu, L., Bai, D., Qiu, W., Liu, B., Wang, Y., Geiger, A., and Liao, Y.
\newblock Hugs: Holistic urban 3d scene understanding via gaussian splatting.
\newblock In \emph{Proceedings of the IEEE/CVF Conference on Computer Vision and Pattern Recognition (CVPR)}, pp.\  21336--21345, June 2024{\natexlab{a}}.

\bibitem[Zhou et~al.(2018)Zhou, Tucker, Flynn, Fyffe, and Snavely]{realestate10K}
Zhou, T., Tucker, R., Flynn, J., Fyffe, G., and Snavely, N.
\newblock Stereo magnification: Learning view synthesis using multiplane images.
\newblock \emph{ACM Trans. Graph. (Proc. SIGGRAPH)}, 37, 2018.

\bibitem[Zhou et~al.(2024{\natexlab{b}})Zhou, Lin, Shan, Wang, Sun, and Yang]{zhou2024drivinggaussian}
Zhou, X., Lin, Z., Shan, X., Wang, Y., Sun, D., and Yang, M.-H.
\newblock Drivinggaussian: Composite gaussian splatting for surrounding dynamic autonomous driving scenes.
\newblock In \emph{Proceedings of the IEEE/CVF Conference on Computer Vision and Pattern Recognition}, pp.\  21634--21643, 2024{\natexlab{b}}.

\bibitem[Zou et~al.(2024)Zou, Yu, Guo, Li, Liang, Cao, and Zhang]{TriplaneGaussian_Zou_2024_CVPR}
Zou, Z.-X., Yu, Z., Guo, Y.-C., Li, Y., Liang, D., Cao, Y.-P., and Zhang, S.-H.
\newblock Triplane meets gaussian splatting: Fast and generalizable single-view 3d reconstruction with transformers.
\newblock In \emph{Proceedings of the IEEE/CVF Conference on Computer Vision and Pattern Recognition (CVPR)}, pp.\  10324--10335, June 2024.

\end{thebibliography}
\bibliographystyle{icml2025}


\end{document}